\documentclass{article}
\usepackage[utf8]{inputenc}
\usepackage{array}
\usepackage{mathtools}
\usepackage{multirow}
\usepackage{amsmath}
\usepackage{amsthm}
\usepackage[pdftex]{graphicx}
\usepackage[numbers,square]{natbib}
\usepackage{microtype}
\usepackage[unicode=true,
 bookmarks=false,
 breaklinks=false,pdfborder={0 0 1},backref=section,colorlinks=false]
 {hyperref}
\hypersetup{
 hidelinks}

\makeatletter

\providecommand{\tabularnewline}{\\}

\usepackage[preprint]{neurips_2018}

\usepackage{times}
\usepackage{soul}
\usepackage[small]{caption}
\usepackage{amsthm}
\usepackage{algorithm}
\usepackage{algorithmic}
\usepackage{amsfonts}

\author{%
  Long Hoang Dang, Thao Minh Le, Vuong Le, Truyen Tran \\
  Applied Artificial Intelligence Institute, Deakin University, Australia\\
  \texttt{\{hldang,lethao,vuong.le,truyen.tran\}@deakin.edu.au} \\
}

\makeatother

\begin{document}
\title{Hierarchical Object-oriented Spatio-Temporal Reasoning for Video Question
Answering}

\maketitle
\global\long\def\vb{\boldsymbol{v}}%
\global\long\def\qb{\boldsymbol{q}}%
\global\long\def\cb{\boldsymbol{c}}%
\global\long\def\rb{\boldsymbol{r}}%
\global\long\def\softmax{\mathrm{softmax}}%
\global\long\def\sigmoid{\mathrm{sigmoid}}%
\global\long\def\softplus{\mathrm{softplus}}%
\global\long\def\ModelName{\text{HOSTR}}%
\global\long\def\UnitName{\text{OSTR}}%

\begin{abstract}
Video Question Answering (Video QA) is a powerful testbed to develop
new AI capabilities. This task necessitates learning to reason about
objects, relations, and events across visual and linguistic domains
in space-time. High-level reasoning demands lifting from associative
visual pattern recognition to symbol-like manipulation over objects,
their behavior and interactions. Toward reaching this goal we propose
an object-oriented reasoning approach in that video is abstracted
as a dynamic stream of interacting objects. At each stage of the video
event flow, these objects interact with each other, and their interactions
are reasoned about with respect to the query and under the overall
context of a video. This mechanism is materialized into a family of
general-purpose neural units and their multi-level architecture called
Hierarchical Object-oriented Spatio-Temporal Reasoning (HOSTR) networks.
This neural model maintains the objects' consistent lifelines in the
form of a hierarchically nested spatio-temporal graph. Within this
graph, the dynamic interactive object-oriented representations are
built up along the video sequence, hierarchically abstracted in a
bottom-up manner, and converge toward the key information for the
correct answer. The method is evaluated on multiple major Video QA
datasets and establishes new state-of-the-arts in these tasks. Analysis
into the model's behavior indicates that object-oriented reasoning
is a reliable, interpretable and efficient approach to Video QA. 
 
\end{abstract}

\section{Introduction}

\begin{figure}
\noindent %
\noindent\begin{minipage}[c]{1\textwidth}%
\begin{minipage}[c]{0.46\textwidth}%
\centering \includegraphics[width=1\columnwidth]{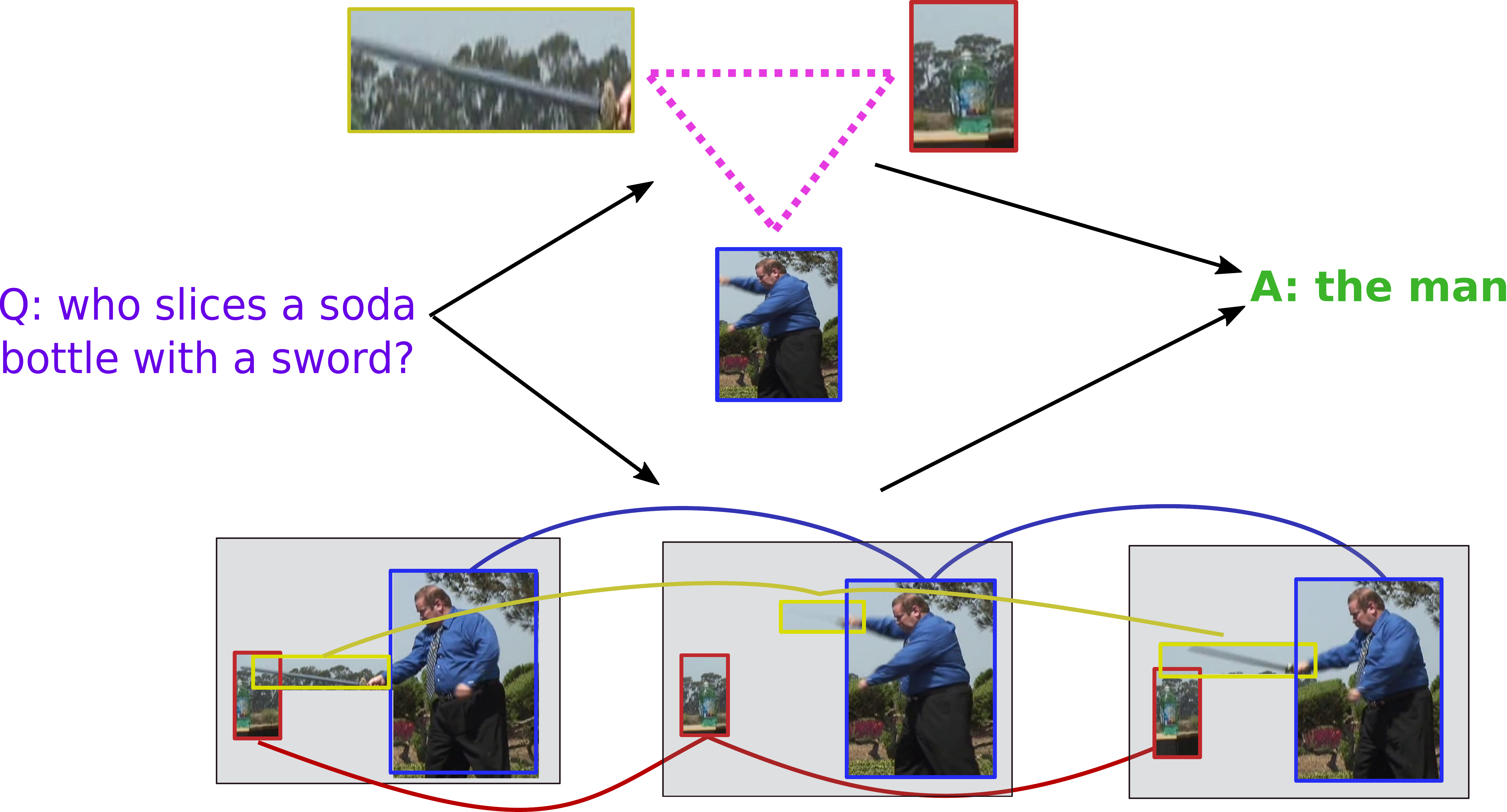}
\captionof{figure}{The key to Video QA is the effective relational
reasoning on objects with their temporal lifelines (below sequences)
interleaved with spatial interactions (upper graph) under the context
set in the video and the perspective provided by the query. This object-oriented
spatio-temporal reasoning is the main theme of this work. \label{fig:teaser}}%
\end{minipage}\hfill{}%
\begin{minipage}[c]{0.46\textwidth}%
\centering{}\includegraphics[width=1\columnwidth]{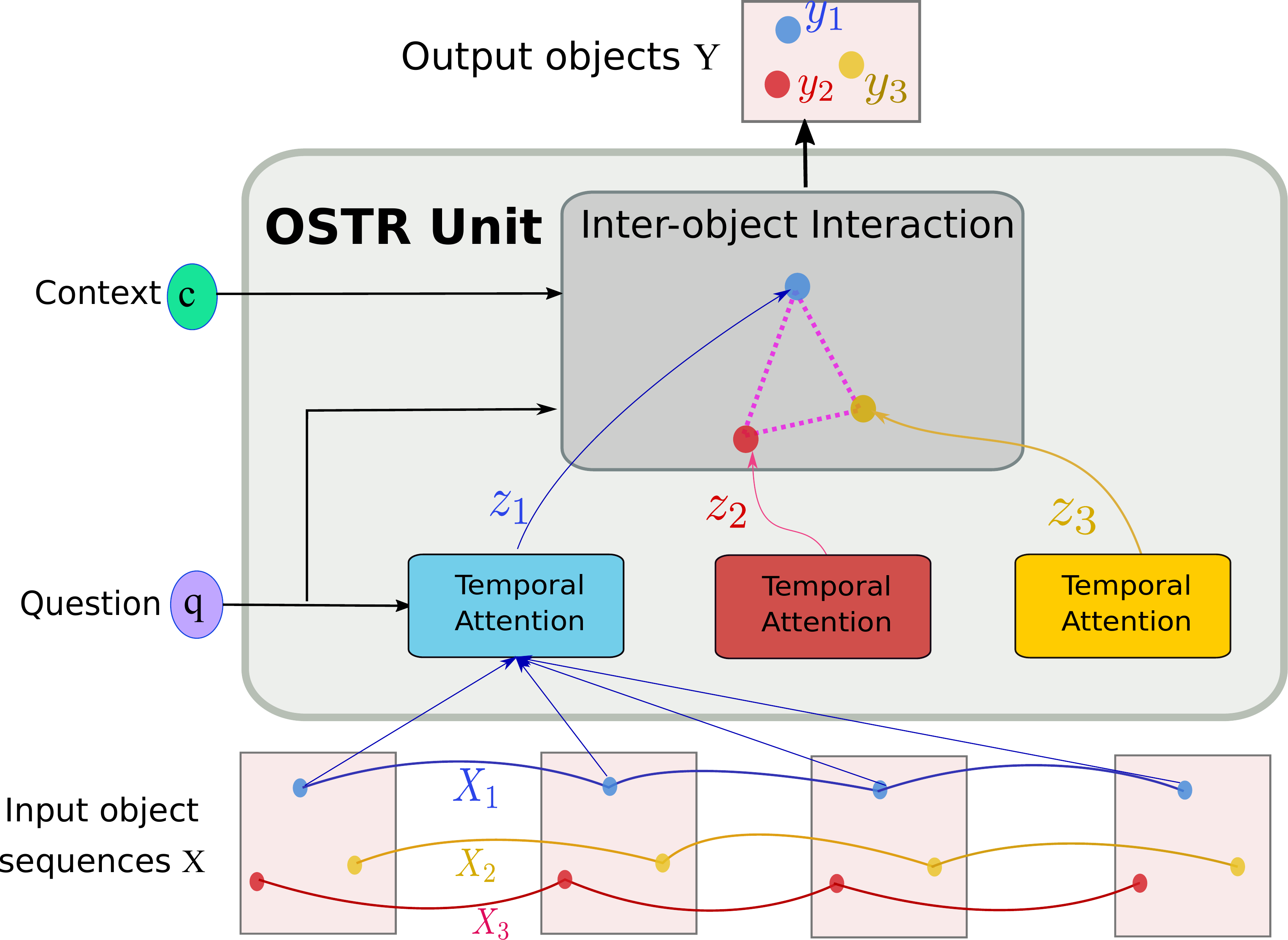}\smallskip{}
 \captionof{figure}{Object-oriented Spatio-Temporal Reasoning
(OSTR) unit. Inputs include a set of object sequences $X$ (with identity
indicated by colors), a context $c$ and a query $q$. Each object
sequence is first summarized by a temporal attention module (matching
color boxes). The inter-object S-T relations is modeled by a graph-based
spatial interaction (gray box with pink graph). OSTR unit outputs
a set of object instances $\left\{ y_{n}\right\} $ with object IDs
corresponding to those in the input sequences. \label{fig:unit}}%
\end{minipage}\vspace{-0.3cm}
\end{minipage}
\end{figure}

Much of the recent impressive progress of AI can be attributed to
the availability of suitable large-scale testbeds. A powerful testbed
-- largely under-explored -- is Video Question Answering (Video
QA). This task demands a wide range of cognitive capabilities including
learning and reasoning about objects and dynamic relations in space-time,
in both visual and linguistic domains. A major challenge of reasoning
over video is extracting question-relevant high-level facts from low-level
moving pixels over an extended period of time. These facts include
objects, their motion profiles, actions, interactions, events, and
consequences distributed in space-time. Another challenge is to learn
the long-term temporal relation of visual objects conditioning on
the guidance clues from the question -- effectively bridging the
semantic gulf between the two domains. Finally, learning to reason
from relational data is an open problem on its own, as it pushes the
boundary of learning from simple one-step classification to dynamically
construct question-specific computational graphs that realize the
iterative reasoning process.

A highly plausible path to tackle these challenges is via \emph{object-centric
learning} since objects are fundamental to cognition \cite{spelke2007core}.
Objects pave the way towards more human-like reasoning capability
and symbolic computing \cite{lake2017building,spelke2007core}. Unlike
objects in static images, objects in video have unique evolving lives
throughout space-time. As object lives throughout the video, it changes
its appearance and position, and interacts with other objects at arbitrary
time. When observed in the videos, all these behaviors play out on
top of a background of rich context of the video scene. Furthermore,
in the question answering setting, these object-oriented information
must be considered from the specific view point set by the linguistic
query. With these principles, we pinpoint the key to Video QA to be
the effective high-level relational reasoning of spatio-temporal objects
under the video context and the perspective provided by the query
(see Fig.~\ref{fig:teaser}). This is challenging for the complexity
of the video spatio-temporal structure and the cross-domain compatibility
gap between linguistic query and visual objects.

Toward such challenge we design a general-purpose neural unit called
Object-oriented Spatio-Temporal Reasoning (OSTR) that operates on
a set of video object sequences, a contextual video feature, and an
external linguistic query. OSTR models object lifelong interactions
and returns a summary representation in the form of a singular set
of objects. The specialties of OSTR are in the partitioning of intra-object
temporal aggregation and inter-object spatial interaction that leads
to the efficiency of the reasoning process. Being flexible and generic,
OSTR units are suitable building blocks for constructing powerful
reasoning models.

For Video QA problem, we use OSTR units to build up Hierarchical Object-oriented
Spatio-Temporal Reasoning (HOSTR) model. The network consists of OSTR
units arranged in layers corresponding to the levels of video temporal
structure. At each level, HOSTR finds local object interactions and
summarizes them toward a higher-level, longer-term representation
with the guidance of the linguistic query.

HOSTR stands out with its authentic and explicit modeling of video
objects leading to the effective and interpretable reasoning process.
The hierarchical architecture also allows the model to efficiently
scale to a wider range of video formats and lengths. These advantages
are demonstrated in a comprehensive set of experiments on multiple
major Video QA tasks and datasets.

In summary, this paper makes three major contributions: (1) A semantic-rich
object-oriented representation of videos that paves the way for spatio-temporal
reasoning (2) A general-purpose neural reasoning unit with dynamic
object interactions per context and query; and (3) A hierarchical
network that produces reliable and interpretable video question answering.

\section{Related Work}

\textbf{Video QA} has been developed on top of traditional video analysis
schemes such as recurrent networks of frame features \cite{zhao2019long}
or 3D convolutional operators \cite{tran2018closer}. Video representations
are then fused with or gated by the linguistic query through co-attention
\cite{jang2017tgif,ye2017video}, hierarchical attention \cite{liang2018focal,zhao2018multi},
and memory networks \cite{kim2017deepstory,wang2019holistic}. More
recent works advance the field by exploiting hierarchical video structure
\cite{le2020hierarchical} or separate reasoning out of representation
learning \cite{le2020neural-reason}. A share feature between these
works is considering the whole video frames or segments as the unit
component of reasoning. In contrast, our work make a step further
by using detail objects from the video as primitive constructs for
reasoning.

\textbf{Object-centric Video Representation} inherits the modern capability
of object detection on images \cite{desta2018object} and continuous
tracking through temporal consistency \cite{wojke2017simple}. Tracked
objects form tubelets \cite{kalogeiton2017action} whose representation
contributes to breakthroughs in action detection \cite{xie2018rethinking}
and event segmentation \cite{chao2018rethinking}. For tasks that
require abstract reasoning, the connection between objects beyond
temporal object permanence can be established through relation networks
\cite{baradel2018object}. The concurrence of objects' 2D spatial-
and 1D temporal- relations naturally forms a 3D spatio-temporal graphs
\cite{wang2018videos}. This graph can be represented as either a
single flattened one where all parts connect together \cite{zeng2019graph},
or separated spatial- and temporal-graphs \cite{pan2020spatio}. They
can also be approximated as a dynamic graph where objects live through
the temporal axis of the video while their properties and connection
evolve \cite{jain2016structural}.

Object-based Video QA is still in infancy. The works in \cite{yang2020bert}
and \cite{huang2020location} extract object features and feed them
to generic relational engines without prior structure of reasoning
through space-time. At the other extreme, detected objects are used
to scaffold the symbolic reasoning computation graph \cite{CLEVRER2020ICLR}
which is explicit but limited in flexibility and cannot recover from
object extraction errors. Our work is a major step toward the object-centric
reasoning with the balance between explicitness and flexibility. Here
video objects serve as active agents which build up and adjust their
interactions dynamically in the spatio-temporal space as instructed
by the linguistic query.

\section{Method}

\subsection{Problem Definition}

Given a video $V$ and linguistic question $q$, our goal is to learn
a mapping function $F_{\theta}(.)$ that returns a correct answer
$\bar{a}$ from an answer set $A$ as follows: 
\begin{equation}
\bar{a}=\arg\max_{a\in A}F_{\theta}\left(a\mid q,\mathcal{V}\right).\label{eq:VQA-overall}
\end{equation}

In this paper, a video $V$ is abstracted as a collection of object
sequences tracked in space and time. Function $F_{\theta}$ is designed
to have the form of a hierarchical object-oriented network that takes
the object sequences, modulates them with the overall video context,
dynamically infers object interactions as instructed by the question
$q$ so that key information regarding $a$ arises from the mix. This
object-oriented representation is presented next in Sec.~\ref{subsec:Linguistic-and-Visual},
followed by Sec.~\ref{subsec:OSTR} describing the key computation
unit and Sec.~\ref{subsec:Hierarchical-Spatio-Temporal-Graph} the
resulting model.

\subsection{Data Representation \label{subsec:Linguistic-and-Visual}}

\subsubsection{Video as a Set of Object Sequences}

Different from most of the prominent VideoQA methods \cite{jang2017tgif,gao2018motion,le2020hierarchical}
where videos are represented by frame features, we break down a video
of length $L$ into a list of $N$ object sequences $O=\left\{ o_{n,t}\right\} _{n=1,t=1}^{N,L}$
constructed by chaining corresponding objects of the same identity
$n$ across the video. These objects are represented by a combination
of (1) appearance features (RoI pooling features) $o_{n,t}^{a}\in\mathbb{R}^{2048}$
(representing ``what''); and (2) the positional features $o_{n,t}^{p}=[\frac{x_{min}}{W},\frac{y_{min}}{H},\frac{x_{max}}{W},\frac{y_{max}}{H},\frac{w}{W},\frac{h}{H},\frac{wh}{WH}]$
(``where''), with $w$ and $h$ are the sizes of the bounding box
and $W,H$ are those of the video frame, respectively.

In practice, we use Faster R-CNN with ROI-pooling to extract positional
and appearance features; and DeepSort for multi-object tracking. We
assume that the objects live from the beginning to the end of the
video. Occluded or missed objects have their features marked with
special null values which will be specially dealt with by the model.

\subsubsection{Joint Encoding of ``What'' and ``Where''}

Since the appearance $o_{n,t}^{a}$ of an object may remain relatively
stable over time while $o_{n,t}^{p}$ constantly changes, we must
find \emph{joint positional-appearance features} of objects to make
them discriminative in both space and time. Specifically, we propose
the following multiplicative gating mechanism to construct such features:
\begin{align}
o_{n,t} & =f_{1}(o_{n,t}^{a})\odot f_{2}(o_{n,t}^{p})\in\mathbb{R}^{d},\label{eq:pos-app-feat}
\end{align}
where $f_{2}(o_{n,t}^{p})\in(\boldsymbol{0},\boldsymbol{1})$ serves
as a position gate to (softly) turn on/off the localized appearance
features $f_{1}(o_{n,t}^{a})$. We choose $f_{1}(x)=\tanh\left(W_{a}x+b_{a}\right)$
and $f_{2}(x)=\text{sigmoid}\left(W_{p}x+b_{p}\right)$, where $W_{a}$
and $W_{p}$ are network weights with height of $d$.

Along with the object sequences, we also maintain global features
of video frames which hold the information of the background scene
and possible missed objects. Specifically, for each frame $t$, we
form the the global\textbf{ }features\textbf{ $g_{t}$} as the combination
of the frame's appearance features (pretrained ResNet $pool5$ vectors)
and motion feature (pretrained ResNeXt-101) extracted from such frame.

With these ready, the video is represented as a tuple of object sequences
$O_{n}$ and frame-wise global features $g_{t}$: $\mathcal{V}=\left(\left\{ O_{n}\mid O_{n}\in\mathbb{R}^{L\times d}\right\} {}_{n=1}^{N},\left\{ g_{t}\right\} _{t=1}^{L}\right)$.

\subsubsection{Linguistic Representation}

We utilize a BiLSTM running on top of GloVe embedding of the words
in a query of length $S$ to generate contextual embeddings $\{e_{s}\}_{s=1}^{S}$
for $e_{s}\in\mathbb{R}^{d}$, which share the dimension $d$ with
object features. We also maintain a global representation of the question
by summarizing the two end LSTM states $q_{g}\in\mathbb{R}^{d}$.
We further use $q_{g}$ to drive the attention mechanism and combine
contextual words into a unified query representation $q=\sum_{s=1}^{S}\alpha_{s}e_{s}$
where $\alpha_{s}=\text{softmax}_{s}(W_{q}(e_{s}\odot q_{g}))$.

\subsection{Object-oriented Spatio-Temporal Reasoning (OSTR) \label{subsec:OSTR}}

\begin{figure*}
\begin{centering}
\includegraphics[width=0.6\paperwidth]{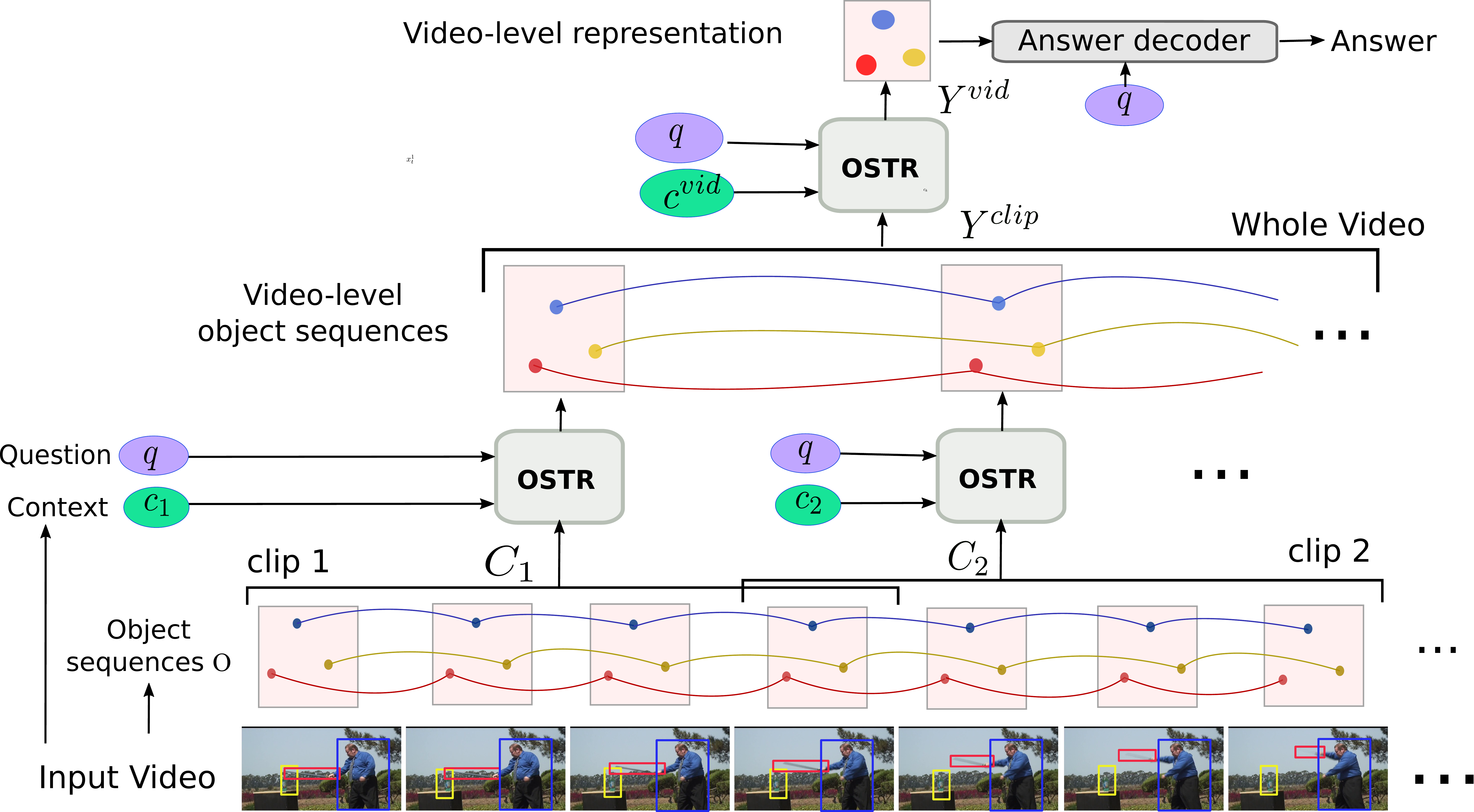} 
\par\end{centering}
\centering{}\caption{The architecture of Hierarchical Object-oriented Spatio-Temporal Reasoning
(HOSTR) network for Video QA. HOSTR contains OSTR units operating
at two levels: clip-level and video-level. Clip-level OSTR units model
the interaction between object chunks within a particular clip under
the modulation of the question and a clip-specific context representation.
Their output objects are chained together and further sent to a video-level
OSTR to capture the long-term dependencies between the objects existing
in the whole video. Finally, a classifier taking as input the summarized
output of video-level OSTR and the query is used for answer prediction.\label{fig:overview}}
\end{figure*}

With the videos represented as object sequences, we need to design
a scalable reasoning framework that can work natively on the structures.
Such a framework must be modular so it is flexible to different input
formats and sizes. Toward this goal, we design a generic reasoning
unit called \emph{Object-oriented Spatio-Temporal Reasoning ($\UnitName$)}
that operates on this object-oriented structure and supports layering
and parallelism.

Algorithmically, OSTR takes as input a query representation $q$,
a context representation $c$, a set of $N$ object sequences $X=\{X_{n}\mid X_{n}\in\mathbb{R}^{T\times d}\}_{n=1}^{N}$
of equal length $T$, and individual identities $\left\{ n\right\} $.
In practice, $X$ can be a subsegment of the whole object sequences
$O$, and $c$ is gathered from the frame features $g_{t}$ constructed
in Sec.~\ref{subsec:Linguistic-and-Visual}. The output of the $\UnitName$
is a set of object instances of the same identity.

Across the space-time domains, real-world objects have distinctive
properties (appearance, position, etc.) and behaviors (motion, deformation,
etc.) throughout their lives. Meanwhile, different objects living
in the same period can interact with each other. The $\UnitName$
closely reflects this nature by containing the two main components:
(1) Intra-object temporal attention and (2) Inter-object interaction
(see Fig.~\ref{fig:unit}).

\subsubsection{Intra-object Temporal Attention \label{par:Temporal-aggregation}}

The goal of the temporal attention module is to produce a \emph{query-specific}
summary of each object sequence $X_{n}=\left\{ x_{n,t}\right\} _{t=1}^{T}$
into a single vector $z_{n}$. The attention weights are driven by
the query $q$ to reflect the fact that the relevance to the query
varies across the sequence. In details, the summarized vector $z_{n}$
is calculated by 
\begin{align}
z_{n} & =\textrm{temporal\_attention}(X_{n})\coloneqq\gamma*\sum_{t=1}^{T}\beta_{t}x_{n,t},\textrm{where}\label{eq:temporal_attention}\\
\beta_{t} & =\text{softmax}_{t}\left(W_{a}\left((W_{q}q+b_{q})\odot(W_{x}x_{t}+b_{x})\right)\right),\label{eq:object_summarized}
\end{align}
where $\odot$ is the Hadamard product, $\left\{ W_{a},W_{q},W_{x}\right\} $
are learnable weights, $\gamma$ is a binary mask vector to handle
the null values caused by missed detections as mentioned in Sec.~\ref{subsec:Linguistic-and-Visual}.

When the sequential structure is particularly strong, we can optionally
employ a BiLSTM to model the sequence. We can then either utilize
the last state of the forward LSTM and the first state of the backward
LSTM, or place attention across the hidden states instead of object
feature embeddings.

\subsubsection{Inter-object Interaction \label{par:Spatial-Interaction}}

Fundamentally, the lifelines of object sequences are not only described
by their internal behavior through time but also by the interactions
with their neighbor objects coexisting in the same space. In order
to represent such complex relationship, we build a spatio-temporal
computation graph to facilitate the inter-object interactions modulated
by the query $q$. This graph $\mathcal{G}(Z,E)$ contains vertices
as the summarized objects $Z=\{z_{n}\}_{n=1}^{N}$ generated in Eq.~\ref{eq:object_summarized},
and the edges $E$ represented by an adjacency matrix $A\in\mathbb{R}^{N\times N}$.
A is calculated dynamically as the \emph{query-induced correlation
matrix }between the objects: 
\begin{align}
a_{n} & =\text{norm}\left(W_{a}([z_{n},z_{n}\odot q])\right),\\
A & =a^{\top}a.\label{eq:adj-mat}
\end{align}
Here $a_{n}$ is the relevance of object $n$ w.r.t. the query $q$.
The \emph{norm} operator is implemented as a softmax function over
objects in our implementation.

Given the graph $\mathcal{G},$we use a Graph Convolutional Network
(GCN) equipped with skip-connections to refine objects in relation
with their neighboring nodes. Starting with the initialization $H^{0}=\left(z_{1},z_{2},...,z_{n}\right)\in\mathbb{R}^{N\times d}$,
the representations of nodes are updated through a number of refinement
iterations. At iteration $i$, the new hidden states are calculated
by:

\begin{align}
\textrm{GCN}_{i}\left(H^{i-1}\right) & =W_{2}^{i-1}\sigma\left(AH^{i-1}W_{1}^{i-1}+b^{i-1}\right)\nonumber \\
H^{i} & =\sigma\left(H^{i-1}+\textrm{GCN}_{i}\left(H^{i-1}\right)\right),\label{eq:GCN}
\end{align}
where $\sigma\left(\cdot\right)$ is a nonlinear activation (ELU in
our implementation). After a fixed number of GCN iterations, the hidden
states of the final layer are gathered as $H^{i_{max}}=\left\{ h_{n}\right\} _{n=1}^{N}$.

To recover the underlying background scene information and compensate
for possible undetected objects, we augment the object representations
with the global context $c$:

\begin{equation}
y_{n}=\textrm{MLP}\left(\left[h_{n};c\right]\right).
\end{equation}
These vectors line up to form the final output of the OSTR unit as
a set of objects $Y=\left\{ y_{n}\right\} _{n=1}^{N}.$

\subsection{Hierarchical Object-oriented Spatio-Temporal Reasoning (HOSTR)\label{subsec:Hierarchical-Spatio-Temporal-Graph}}

Even though the partitioning of temporal and spatial interaction in
$\UnitName$ brings the benefits of efficiency and modularity, such
separated treatment can cause the loss of spatio-temporal information,
especially with long sequences. This limitation prevents us from using
OSTR directly on the full video object sequences. To allow temporal
and spatial reasonings to cooperate along the way, we break down a
long video into multiple short (overlapping) clips and impose an hierarchical
structure on top. With such division, the two types of interactions
can be combined and interleaved across clips and allow full spatio-temrporal
reasoning.

Based on this motive, we design a novel hierarchical structure called
Hierarchical Object-oriented Spatio-Temporal Reasoning ($\ModelName$)
that follows the video multi-level structure and utilizes the $\UnitName$
units as building blocks. Our architecture shares the design philosophy
of hierarchical reasoning structures with HCRN \cite{le2020hierarchical}
as well as the other general neural building blocks such as ResNet
and InceptionNet. Thanks to the genericity of $\UnitName$, we \emph{can
build a hierachy of arbitrary depth}. For concreteness, we present
here a two-layer HOSTR corresponding to the video structure: clip-level
and video-level (see Fig.~\ref{fig:overview}).

In particular, we first split all object sequences $O$ constructed
in Sec.~\ref{subsec:Linguistic-and-Visual} into $K$ equal-sized
chunks $C=\left\{ C_{1},C_{2},...,C_{K}\right\} $ corresponding to
the video clips, each of $T$ frames. As the result, each chunk includes
the object subsequences $C_{k}=\left\{ o_{n,t}\right\} _{n=1,t=t_{k}}^{N,t_{k}+T}$,
where $o_{n,t}$ are the object features extracted in Eq.~\ref{eq:pos-app-feat},
and $t_{k}$ is the starting time of clip $k$.

Similarly, we divide the sequence of global frame features $\left\{ g_{t}\right\} _{t=1}^{L}$
into $K$ parts corresponding to the video clips. The global context
$c_{k}$ for clip $k$ is derived from each part by an identical operation
with the temporal attention for objects in Eqs.~\ref{eq:temporal_attention},\ref{eq:object_summarized}:
$c_{k}=\textrm{temporal\_attention}\left(\left\{ g_{t}\right\} _{t=t_{k}}^{t_{k}+T}\right)$.

Clip-level OSTR units work on each of these subsequences $C_{k}$,
context $c_{k}^{\textrm{clip}}$and query $q$, and generate the clip-level
representation of the chunk $y_{k}^{\textrm{clip}}\in R^{N\times d}$:

\begin{equation}
y_{k}^{\textrm{clip}}=\textrm{OSTR}(C_{k},c_{k}^{\textrm{clip}},q).\label{eq:clip-ostr}
\end{equation}
Outputs of the $K$ clip-level OSTRs are $K$ different sets of objects
$y_{k}^{\textrm{clip}}=\left\{ y_{n,k}\right\} _{n=1}^{N}$ whose
identities $n$ were maintained. Therefore, we can easily chain these
objects of the same identity from different clips together to form
the video-level sequence of objects $Y^{\textrm{clip}}=\left\{ y_{n,k}^{\textrm{clip}}\right\} _{n=1,k=1}^{N,K}$.

At the video level, we have a single OSTR unit that takes in the object
sequence $Y^{\textrm{clip}}$, query $q,$ and video-level context
$c^{\textrm{vid}}$. The context $c^{\textrm{vid}}$ is again derived
from the clip-level context $c_{k}^{\textrm{clip}}$ by temporal attention:
$c^{vid}=\textrm{temporal\_attention}\left(\left\{ c_{k}^{\textrm{clip}}\right\} _{k=1}^{K}\right)$
.

The video-level $\UnitName$ models the long-term relationships between
input object sequences in the whole video: 
\[
Y^{vid}=\textrm{OSTR}(Y^{\textrm{clip}},c^{\textrm{vid}},q).
\]
The output of this unit is a set of $N$ vectors $Y^{\textrm{vid}}=\{y_{n}^{\textrm{vid}}\mid y_{n}^{\textrm{vid}}\in\mathbb{R}^{d}\}_{n=1}^{N}$.
The set is further summarized using an attention mechanism using the
query $q$ into the final representation vector $r$:

\vspace{-3mm}

\begin{align}
\delta_{n} & =\text{softmax}_{n}\left(\textrm{MLP}\left[W_{y}y_{n}^{\textrm{vid}};W_{y}y_{n}^{\textrm{vid}}\odot W_{c}q\right]\right),\\
r & =\sum_{n=1}^{N}\delta_{n}y_{n}^{vid}\in\mathbb{R}^{d}.
\end{align}

\subsection{Answer Decoders}

We follow the common settings for answer decoders (e.g., see \cite{jang2017tgif})
which combine the final representation $r$ with the query $q$ using
an MLP followed by a softmax to rank the possible answer choices.
More details about the answer decoders per question types are available
in the supplemental material. We use the cross-entropy as the loss
function to training the model from end to end for all tasks except
counting, where Mean Square Error is used.

\section{Experiments}

\subsection{Datasets \label{subsec:Datasets}}

We evaluate our proposed $\ModelName$ on the three public video QA
benchmarks, namely, TGIF-QA \cite{jang2017tgif}, MSVD-QA \cite{xu2017video}
and MSRVTT-QA \cite{xu2017video}. More details are as follows.

\textbf{MSVD-QA} consists of 50,505 QA pairs annotated from 1,970
short video clips. The dataset covers five question types: \emph{What},
\emph{Who}, \emph{How}, \emph{When}, and \emph{Where}, of which 61\%
of the QA pairs for training, 13\% for validation and 26\% for testing.

\textbf{MSRVTT-QA} contains 10K real videos (65\% for training, 5\%
for validation, and 30\% for testing) with more than 243K question-answer
pairs. Similar to MSVD-QA, questions are of five types: \emph{What},
\emph{Who}, \emph{How}, \emph{When}, and \emph{Where}.

\textbf{TGIF-QA} is one of the largest Video QA datasets with 72K
animated GIFs and 120K question-answer pairs. Questions cover four
tasks\emph{ - Action, Event Transition, FrameQA, Count}. We refer
readers to the supplemental material for the dataset description and
statistics.

\subsection{Comparison Against SOTAs}

\textbf{Implementation:} We use Faster R-CNN\footnote{https://github.com/airsplay/py-bottom-up-attention}
for frame-wise object detection. The number of object sequences per
video for MSVD-QA , MSRVTT-QA is 40 and TGIF-QA is 50. We embed question
words into 300-D vectors and initialize them with GloVe during training.
Default settings are with 6 GCN layers for each $\UnitName$ unit.
The feature dimension $d$ is set to be $512$ in all sub-networks.

We compare the performance of $\ModelName$ against recent state-of-the-art
(SOTA) methods on all three datasets. Prior results are taken from
\cite{le2020hierarchical}.

\textbf{MSVD-QA and MSRVTT-QA}: Table~\ref{tab:SOTA-MSVD} shows
detailed comparisons on MSVD-QA and MSRVTT-QA datasets. It is clear
that our proposed method consistently outperforms all SOTA models.
Specifically, we significantly improve performance on the MSVD-QA
by 3.3 absolute points while the improvement on the MSRVTT-QA is more
modest. As videos in the MSRVTT-QA are much longer (3 times longer
than those in MSVD-QA) and contain more complicated interaction, it
might require a larger number of input object sequences than what
in our experiments (40 object sequences).

\begin{table}
\begin{centering}
\begin{tabular}{l|cc}
\hline 
\multirow{1}{*}{Model} & \multicolumn{2}{c}{Test Accuracy (\%)}\tabularnewline
\hline 
 & MSVD-QA  & MSRVTT-QA\tabularnewline
\hline 
ST-VQA  & 31.3  & 30.9\tabularnewline
Co-Mem  & 31.7  & 32.0\tabularnewline
AMU  & 32.0  & 32.5\tabularnewline
HME  & 33.7  & 33.0\tabularnewline
HCRN  & 36.1  & 35.4\tabularnewline
\hline 
HOSTR  & \textbf{39.4}  & \textbf{35.9}\tabularnewline
\hline 
\end{tabular}
\par\end{centering}
\vspace{0.3cm}

\caption{Experimental results on MSVD-QA and MSRVTT-QA. \label{tab:SOTA-MSVD}}

\vspace{-0.3cm}
\end{table}

\begin{table}
\begin{centering}
\begin{tabular}{l|c|c|c|c}
\hline 
\multirow{1}{*}{Model} & \multicolumn{4}{c}{TGIF-QA}\tabularnewline
\hline 
 & Action$\uparrow$  & Trans.$\uparrow$  & Frame$\uparrow$  & Count$\downarrow$\tabularnewline
\hline 
ST-TP (R+C)  & 62.9  & 69.4  & 49.5  & 4.32\tabularnewline
Co-Mem (R+F)  & 68.2  & 74.3  & 51.5  & 4.10\tabularnewline
PSAC (R)  & 70.4  & 76.9  & 55.7  & 4.27\tabularnewline
HME (R+C)  & 73.9  & 77.8  & 53.8  & 4.02\tabularnewline
HCRN (R)  & 70.8  & 79.8  & 56.4  & 4.38\tabularnewline
HCRN (R+F)  & 75.0  & 81.4  & 55.9  & 3.82\tabularnewline
\hline 
HOSTR (R)  & \textbf{75.6}  & 82.1  & \textbf{58.2}  & 4.13\tabularnewline
HOSTR (R+F)  & 75.0  & \textbf{83.0}  & 58.0  & \textbf{3.65}\tabularnewline
\hline 
\end{tabular}
\par\end{centering}
\vspace{0.3cm}

\caption{Experimental results on TGIF-QA dataset. R: ResNet, F: Flow, and C:
C3D, respectively. MSE is used as the evaluation metric for count
while accuracy is used for others.\label{tab:SOTA-TGIF}}

\vspace{-0.3cm}
\end{table}

\textbf{TGIF-QA:} Table~\ref{tab:SOTA-TGIF} presents the results
on TGIF-QA dataset. As pointed out in \cite{le2020hierarchical},
short-term motion features are helpful for action task while long-term
motion features are crucial for event transition and count tasks.
Hence, we provide two variants of our model: \emph{HOSTR (R)} makes
use of ResNet features as the context information for $\UnitName$
units; and \emph{HOSTR (R+F)} makes use of the combination of ResNet
features and motion features (extracted by ResNeXt, the same as in
HCRN) as the context representation. $\ModelName$ (R+F) shows exceptional
performance on tasks related to motion. Note that we only use the
context modulation at the video level to concentrate on the long-term
motion. Even without the use of motion features, $\ModelName$ (R)
consistently shows more favorable performance than existing works.

The quantitative results prove the effectiveness of the object-oriented
reasoning compared to the prior approaches of totally relying on frame-level
features. Incorporating the motion features as context information
also shows the flexibility of the design, suggesting that HOSTR can
leverage a variety of input features and has the potentials to apply
to other problems.

\subsection{Ablation Studies}

\begin{figure}
\noindent %
\noindent\begin{minipage}[c]{1\textwidth}%
\begin{minipage}[c]{0.46\textwidth}%
\centering %
\begin{tabular}{l|c}
\hline 
\multirow{1}{*}{{\footnotesize{}Model}} & {\footnotesize{}Test Acc. (\%)}\tabularnewline
\hline 
{\footnotesize{}Default config. ({*})} & {\footnotesize{}39.4}\tabularnewline
\hline 
\textbf{\footnotesize{}Temporal attention (TA)} & \tabularnewline
{\footnotesize{}\quad{}Attention at both levels} & {\footnotesize{}39.4}\tabularnewline
{\footnotesize{}\quad{}BiLSTM at clip, TA at video level} & {\footnotesize{}39.4}\tabularnewline
{\footnotesize{}\quad{}BiLSTM at both levels} & {\footnotesize{}38.8}\tabularnewline
\hline 
\textbf{\footnotesize{}Inter-object Interaction} & \tabularnewline
{\footnotesize{}\quad{}SR with 1 GCN layer} & {\footnotesize{}38.3}\tabularnewline
{\footnotesize{}\quad{}SR with 4 GCN layers} & {\footnotesize{}38.7}\tabularnewline
{\footnotesize{}\quad{}SR with 8 GCN layers} & {\footnotesize{}39.0}\tabularnewline
\hline 
\textbf{\footnotesize{}Contextual representation} & \tabularnewline
{\footnotesize{}\quad{}w/o contextual representation} & {\footnotesize{}37.8}\tabularnewline
\hline 
\textbf{\footnotesize{}Hierarchy} & \tabularnewline
{\footnotesize{}\quad{}1-level hierarchy} & {\footnotesize{}38.0}\tabularnewline
{\footnotesize{}\quad{}1.5-level hierarchy} & {\footnotesize{}38.7}\tabularnewline
\hline 
\end{tabular} \captionof{table}{Ablation results on MSVD-QA dataset. Default
config. ({*}): 6 GCN layers, Attention at clip \& BiLSTM at video
level \label{tab:Ablation-results}}%
\end{minipage}\hfill{}%
\begin{minipage}[c]{0.46\textwidth}%
\centering{}\includegraphics[width=0.95\columnwidth]{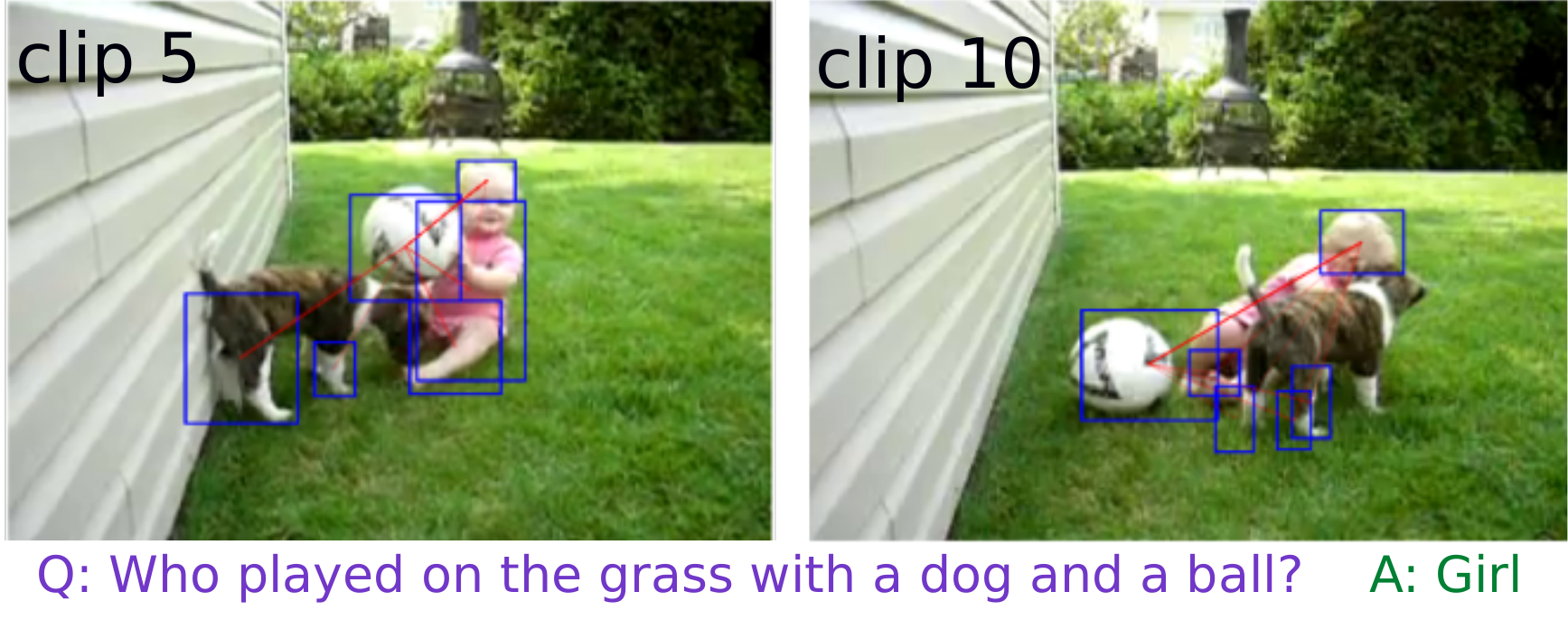}\smallskip{}
 \captionof{figure}{A visualization of the spatio-temporal graph
formed in HOSTR. The six most attended objects in each clip are drawn
in blue bounding boxes. Red links indicates the importances of the
edges. \label{fig:visualization}}%
\end{minipage}%
\end{minipage}
\end{figure}

To provide more insight about our model, we examine the contributions
of different design components to the model's performance on the MSVD-QA
dataset. We detail the results in Table~\ref{tab:Ablation-results}.
As shown, intra-object temporal attention seems to be more effective
in summarizing the input object sequences to work with relational
reasoning than BiLSTM. We hypothesize that it is due to the selective
nature of the attention mechanism -- it keeps only information relevant
to the query.

As for the inter-object interaction, increasing the number of GCN
layers up to 6 generally improves the performance. It gradually degrades
when we stack more layers due to the gradient vanishing. The ablation
study also points out the significance of the contextual representation
as the performance steeply drops from 39.4 to 37.8 without them.

Last but not least, we conduct two experiments to demonstrate the
significance of video hierarchical modeling. \emph{``1-level hierarchy''}
refers to when we replace all clip-level $\UnitName$s with the global
average pooling operation to summarize each object sequence into a
vector while keeping the $\UnitName$ unit at the video level. \emph{``1.5-level
hierarchy''}, on the other hand, refers to when we use an average
pooling operation at the video level while keeping the clip-level
the same as in our $\ModelName$. Empirically, it shows that going
deeper in hierarchy consistently improves performance on this dataset.
The hierarchy may have greater effects in handling longer videos such
as those in the MSRVTT-QA and TGIF-QA datasets.

\subsection{Qualitative Analysis}

To provide more analysis on the behavior of $\ModelName$ in practice,
we visualize the spatio-temporal graph formed during HOSTR operation
on a sample in MSVD-QA dataset. In Fig.~\ref{fig:visualization},
the spatio-temporal graph of the two most important clips (judged
by the temporal attention scores) are visualized in order of their
appearance in the video's timeline. Blue boxes indicate the six objects
with highest edge weights (row summation of the adjacency matrix $A$
calculated in Eq.\ref{eq:adj-mat}). The red lines indicates the most
prominent edges of the graph with intensity scaled to the edge strength.

In this example, HOSTR attended mostly on the objects related to the
concepts relevant to answer the question (the girl, ball and dog).
Furthermore, the relationships between the girl and her surrounding
objects are the most important among the edges, and this intuitively
agrees with how human might visually examine the scene given the question.

\section{Conclusion}

We presented a new object-oriented approach to Video QA where objects
living in the video are treated as the primitive constructs. This
brings us closer to symbolic reasoning, which is arguably more human-like.
To realize this high-level idea, we introduced a general-purpose neural
unit dubbed Object-oriented Spatio-Temporal Reasoning (OSTR). The
unit reasons about its contextualized input -- which is a set of
object sequences -- as instructed by the linguistic query. It first
selectively transforms each sequence to an object node, then dynamically
induces links between the nodes to build a graph. The graph enables
iterative relational reasoning through collective refinement of object
representation, gearing toward reaching an answer to the given query.
The units are then stacked in a hierarchy that reflects the temporal
structure of a typical video, allowing higher-order reasoning across
space and time. Our architecture establishes new state-of-the-arts
on major Video QA datasets designed for complex compositional questions,
relational, temporal reasoning. Our analysis shows that object-oriented
reasoning is a reliable, interpretable and effective approach to Video
QA.

{\small{}\bibliographystyle{abbrvnat}
\bibliography{egbib}
}{\small\par}

\section*{Appendix}

\section{Implementation Details}

\subsection{Question embedding}

Each word in a question length $S$ is first embedded into a vector
space of dimension. The vector sequence is the input of a BiLSTM to provide a sequence of forward-backward states. The hidden state pairs is gathered to
form the contextual word representations $\{e_{s}\}_{s=1}^{S}\ e_{s}\in\mathbb{R}^{d}$,
where $d$ is the length of the vector. The global representation of the
question is then summarized using the two end states: $q_{g}=\left[\overleftarrow{e_{1}};\overrightarrow{e_{S}}\right],\ q_{g}\in\mathbb{R}^{d},$where
$[\thinspace;]$ denotes vector concatenation operation.

\subsection{Answer decoders}

In this subsection, we provide further details on how HOSTR predicts
answers given the final representation $r$ and the query representation
$q$. Depending on question types, we design slightly different answer
decoders. As for open-ended (MSVD-QA, MSRVTT-QA and FrameQA task in
TGIF-QA) and multiple-choice questions (Action and Event Transition
task in TGIF-QA), we utilize a classifier of 2-fully connected layers with the softmax function to rank words in a predefined vocabulary
set $A$ or rank answer choices:
\begin{align}
z & =\text{MLP}(W_{r}\left[r;W_{q}q+b_{q}\right]+b_{r}),\label{eq:regression}\\
p & =\text{softmax}(W_{z}z+b_{z}).\label{eq:class_probs}
\end{align}
We use the cross-entropy as the loss function to training the model
from end to end in this case. While prior works \cite{gao2018motion,le2020hierarchical,jang2017tgif}
rely on hinge loss to train multiple-choice questions, we find that
cross-entropy loss produces more favorable performance. For the count
task in TGIF-QA dataset, we simply take the output of Eq. \ref{eq:regression}
and further feed it into a rounding operation for an integer output
prediction. We use Mean Squared Error as the training loss for this
task.

\subsection{Training}

In this paper, each video is split into $K=10$ clips of $T=10$ consecutive
frames. This is done simply based on the empirical results. For the
HOSTR (R+F) variant, we divided each video into eight clips of $T=16$
consecutive frames to fit in the predefined configuration of the ResNeXt
pretrained model\footnote{https://github.com/kenshohara/video-classification-3d-cnn-pytorch}.
To mitigate the loss of temporal information when split a video up,
there are overlapping parts between two video clips next to each other.

We train our model using Adam optimizer \cite{kingma2014adam} at
an initial learning rate of $10^{-4}$ and weight decay of $10^{-5}$.
The learning rate is reduced after every 10 epochs for all tasks except
the count task in TGIF-QA in which we reduce the learning rate by
half after every 5 epochs. We use a batch size of 64 for MSVD-QA and
MSRVTT-QA while that of TGIF-QA is 32. To be compatible with related
works \cite{le2020neural-reason,emrvqasongMM18,8451103,le2020hierarchical,fan2019heterogeneous},
we use accuracy as evaluation metric for multi-class classification
tasks and MSE for the count task in TGIF-QA dataset.. 

All experiments are conducted on a single GPU NVIDIA Tesla V100-SXM2-32GB
installed on a sever of 40 physical processors and 256 GB of memory
running on Ubuntu 18.0.4. It may require a GPU of at least 16GB in
order to accommodate our implementation. We use early stopping when
validation accuracy decreases after 10 consecutive epochs. Otherwise,
all experiments are terminated after 25 epochs at most. Depending
on dataset sizes, it may take 9 hours (MSVD-QA) to 30 hours (MSRVTT-QA)
of training. As for inference, it takes up to 40 minutes to complete
(on MSRVTT-QA). HOSTR is implemented in Python 3.6 with Pytorch 1.2.0.

\begin{figure*}[t]
\includegraphics[width=1\textwidth]{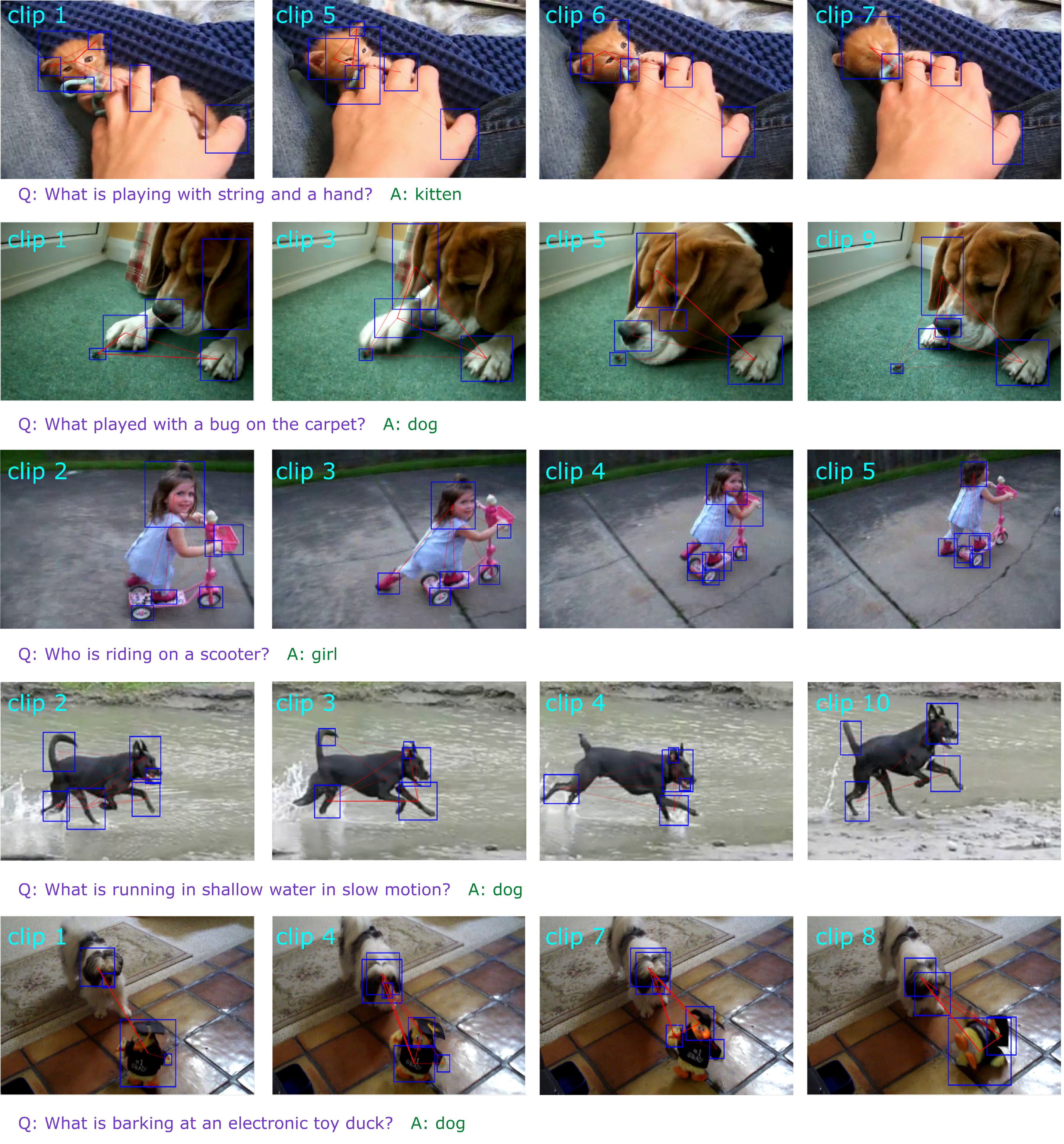}

\caption{Extra visualization of the spatio-temporal graph formed in HOSTR,
similar to Fig.4 in the main manuscript. The most attended objects
in each clip are drawn in blue bounding boxes. Red links indicates
the importances of the edges\label{fig:A-visualization-of}}

\vspace{-0.6cm}
\end{figure*}

\section{Dataset Description}

\begin{table}
\begin{centering}
\begin{tabular}{c|c|c|c|c|c|c|c}
\hline 
\multirow{2}{*}{} & \multirow{2}{*}{Video} & \multirow{2}{*}{Q-A pairs} & \multicolumn{5}{c}{Per question types}\tabularnewline
\cline{4-8} \cline{5-8} \cline{6-8} \cline{7-8} \cline{8-8} 
 &  &  & What & Who & How & When & Where\tabularnewline
\hline 
Train & 1,200 & 30,933 & 19,485 & 10,479 & 736 & 161 & 72\tabularnewline
Val & 250 & 6,415 & 3,995 & 2,168 & 185 & 51 & 16\tabularnewline
Test & 520 & 13,157 & 8,149 & 4,552 & 370 & 58 & 28\tabularnewline
\hline 
All & 1,970 & 50,505 & 31,625 & 17,199 & 1,291 & 270 & 116\tabularnewline
\hline 
\end{tabular}
\par\end{centering}
\vspace{0.3cm}

{\small{}\caption{Statistics of the MSVD-QA dataset.\label{tab:msvd_stats}}
}{\small\par}
\end{table}

\begin{table}
\begin{centering}
\begin{tabular}{c|c|c|c|c|c|c|c}
\hline 
\multirow{2}{*}{} & \multirow{2}{*}{Video} & \multirow{2}{*}{Q-A pairs} & \multicolumn{5}{c}{Per question types}\tabularnewline
\cline{4-8} \cline{5-8} \cline{6-8} \cline{7-8} \cline{8-8} 
 &  &  & What & Who & How & When & Where\tabularnewline
\hline 
Train & 6,513 & 158,581 & 108,792 & 43,592 & 4,067 & 1,626 & 504\tabularnewline
Val & 497 & 12,278 & 8,337 & 3,493 & 344 & 106 & 52\tabularnewline
Test & 2,990 & 72,821 & 49,869 & 20,385 & 1,640 & 677 & 250\tabularnewline
\hline 
All & 10,000 & 243,680 & 166,998 & 67,416 & 6,051 & 2,409 & 806\tabularnewline
\hline 
\end{tabular}
\par\end{centering}
\vspace{0.3cm}

\caption{Statistics of the MSRVTT-QA dataset.\label{tab:msrvtt_stats}}
\end{table}

\begin{table}
\begin{centering}
\begin{tabular}{c|c|c|c|c|c}
\hline 
\multirow{2}{*}{} & \multirow{2}{*}{Q-A pairs} & \multicolumn{4}{c}{Per tasks}\tabularnewline
\cline{3-6} \cline{4-6} \cline{5-6} \cline{6-6} 
 &  & Action & Transition & FrameQA & Count\tabularnewline
\hline 
Train & 1,254,470 & 18,427 & 47,433 & 35,452 & 24,158\tabularnewline
Val & 13,944 & 2,048 & 5,271 & 3,940 & 2,685\tabularnewline
Test & 25,751 & 2,274 & 6,232 & 13,691 & 3,554\tabularnewline
\hline 
All & 1,294,165 & 22,749 & 58,936 & 53,083 & 30,397\tabularnewline
\hline 
\end{tabular}
\par\end{centering}
\vspace{0.3cm}

\caption{Statistics of the TGIF-QA dataset.\label{tab:tgif-qa}}
\end{table}

\paragraph{MSVD-QA}

is a relatively small dataset of 50,505 QA pairs annotated from 1,970
short video clips. The dataset covers five question types: \emph{What},
\emph{Who}, \emph{How}, \emph{When}, and \emph{Where}, of which 61\%
of the QA pairs for training, 13\% for validation and 26\% for testing.
We provide details on the number of question-answer pairs in Table
\ref{tab:msvd_stats}.

\paragraph{MSRVTT-QA}

contains 10K real videos (65\% for training, 5\% for validation, and
30\% for testing) with more than 243K question-answer pairs. Similar
to MSVD-QA, questions are of five types: \emph{What}, \emph{Who},
\emph{How}, \emph{When}, and \emph{Where}. Details on the statistics
of the MSRVTT-QA dataset is shown in Table \ref{tab:msrvtt_stats}.

\paragraph{TGIF-QA}

is one of the largest Video QA datasets of 72K animated GIFs and 120K
question-answer pairs. Questions cover four tasks\emph{ - Action}:
multiple-choice task identifying what action repeatedly happens in
a short period of time; \emph{Transition}: multiple-choice task assessing
if machines could recognize the transition between two events in a
video; \emph{FrameQA: }answers can be found in one of video frames
without the need of temporal reasoning; and \emph{Count}: requires
machines to be able to count the number of times an action taking
place. We take 10\% the number of Q-A pairs and their associated videos
in the training set per each task as the validation set. Details on
the number of Q-A pairs and videos per split are provided in Table
\ref{tab:tgif-qa}.

\section{Qualitative Results}

In addition to the example provided in the main paper, we present
a few more examples of the spatio-temporal graph formed in the HOSTR
model in Figure \ref{fig:A-visualization-of}. In all examples presented,
our HOSTR has successfully modeled the relationships between the targeted
objects in the questions or their parts with surrounding objects.
These examples clearly demonstrate the explainability and transparency
of our model, explaining how the model arrives at the correct answers.

\section{Complexity Analysis}

We analyze the memory consumption on the number of object $n$, hierarchy
depth $h$ given that other factors like video length are fixed; and
OSTR units in one layer share parameters. 

There are two types of memory $m_{a}$ is from hidden unit activations,
$m_{p}$ is for parameters and related terms. We found that $m_{a}$
is linear with \emph{$n$}, constant to $h$. On the other hand, $m_{p}$
is linear with $h$ and constant to $n$. These figures are on-par
with commonly used sequential models. No major extra memory consumption
is introduced.

\end{document}